\documentclass[electronic]{vgtc}             %

\ifpdf%
  \pdfoutput=1\relax                   %
  \usepackage{graphicx}                %
  \DeclareGraphicsExtensions{.pdf,.png,.jpg,.jpeg} %
\else%
  \usepackage{graphicx}                %
  \DeclareGraphicsExtensions{.eps}     %
\fi%

\graphicspath{{figures/}{pictures/}{images/}{./}} %

\usepackage{amssymb}
\usepackage{float}
\usepackage{multirow}
\usepackage{enumitem}

\usepackage{microtype}                 %
\PassOptionsToPackage{warn}{textcomp}  %
\usepackage{textcomp}                  %
\usepackage{mathptmx}                  %
\usepackage{times}                     %
\usepackage{cite}                      %
\usepackage{tabu}                      %
\usepackage{booktabs}                  %

\vgtccategory{Research}

\vgtcinsertpkg

\title{Animatable Virtual Humans: Learning pose-dependent human representations in UV space for interactive performance synthesis}

\author{
Wieland Morgenstern\textsuperscript{1}
\quad Milena T. Bagdasarian\textsuperscript{1,2}
\quad Anna Hilsmann\textsuperscript{1} 
\quad Peter Eisert\textsuperscript{1,2} 
\\[0.2cm]
\textsuperscript{1} Fraunhofer Heinrich Hertz Institute, HHI \\ \textsuperscript{2} Humboldt University of Berlin
\\[0.2cm]
{\tt\small \{first\}.\{last\}@hhi.fraunhofer.de}
}

\teaser{
  \centering
  \includegraphics[width=\linewidth]{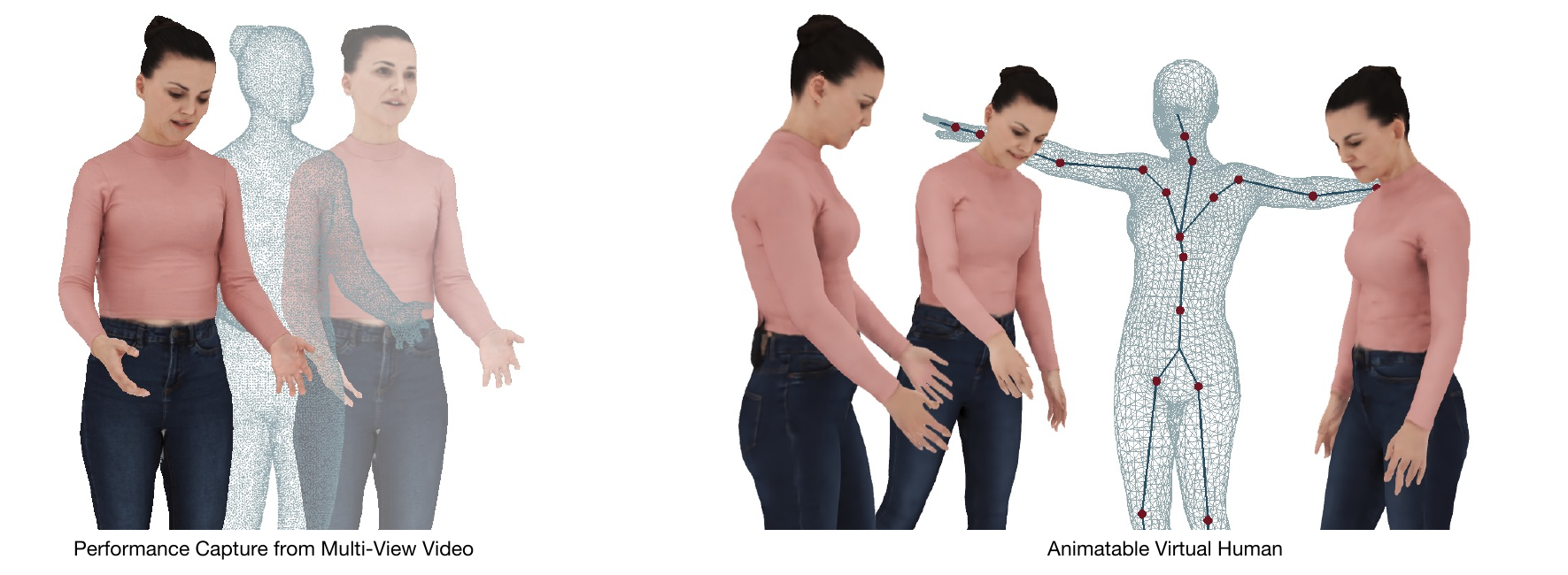}
  \caption{We learn a freely animatable virtual human from a sequence of textured meshes captured in a multi-camera studio.}
  \label{fig:teaser}
}

\abstract{We propose a novel representation of virtual humans for highly realistic real-time animation and rendering in 3D applications. We learn pose dependent appearance and geometry from highly accurate dynamic mesh sequences obtained from state-of-the-art multiview-video reconstruction. Learning pose-dependent appearance and geometry from mesh sequences poses significant challenges, as it requires the network to learn the intricate shape and articulated motion of a human body. However, statistical body models like SMPL provide valuable a-priori knowledge which we leverage in order to constrain the dimension of the search space enabling more efficient and targeted learning and define pose-dependency. Instead of directly learning absolute pose-dependent geometry, we learn the difference between the observed geometry and the fitted SMPL model. This allows us to encode both pose-dependent appearance and geometry in the consistent UV space of the SMPL model. This approach not only ensures a high level of realism but also facilitates streamlined processing and rendering of virtual humans in real-time scenarios.

} %

\begin{document}

\maketitle

\firstsection{Introduction}

\begin{figure*}
    \centering
    \includegraphics[width=\textwidth]{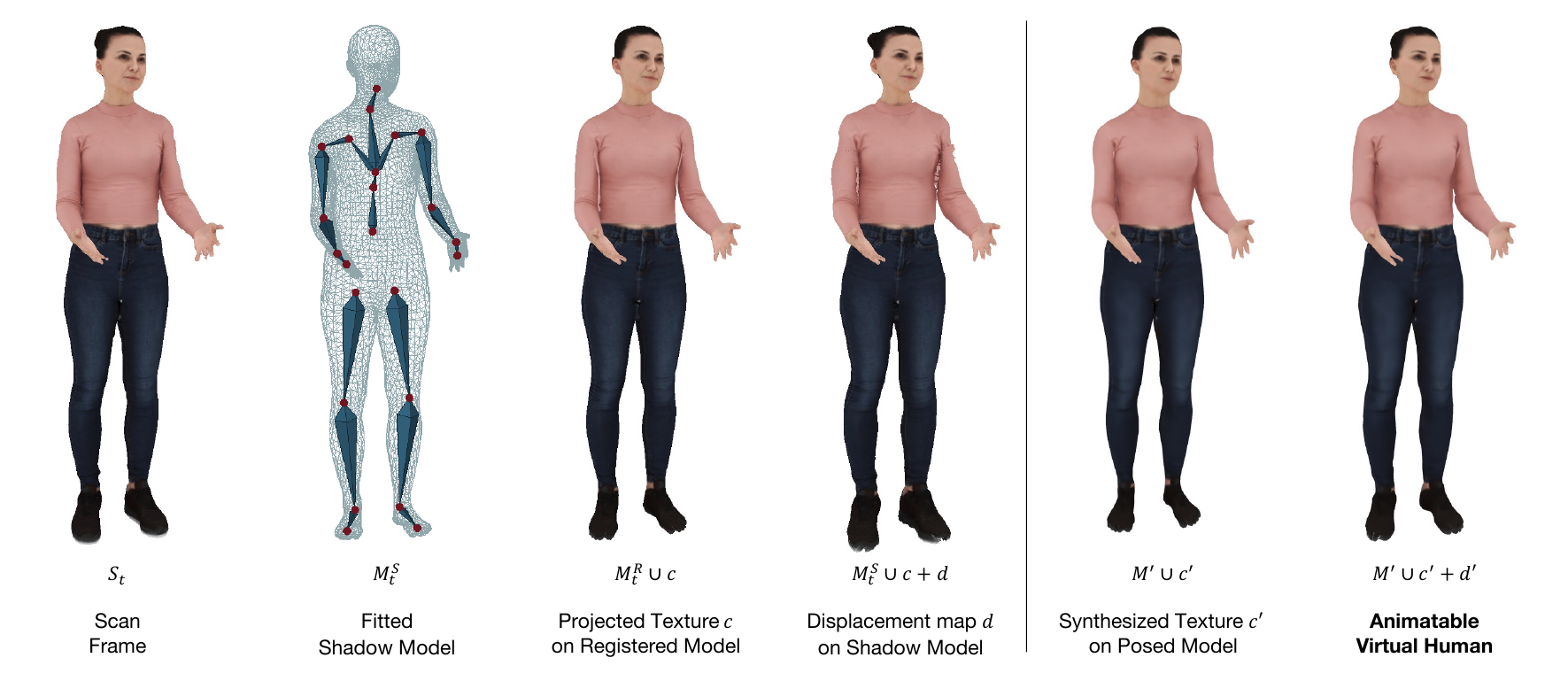}
    \caption{The four images on the left show a reconstructed model from a performance capture in a fixed pose ($S_t$), the shadow model ($M_t^S$), the registered ($M_t^R$) model with the projected texture and finally the added displacement map. This process leaves artifacts in the texture $c$, and the computed displacement $d$ is noisy when re-applied. The two images on the right depict the animatable representation with texture $c'$ and displacement $d'$, which we can synthesize in novel poses. Here, rendered in the same pose as the captured frame to allow for direct visual comparison. It is apparent that the neural network has learned to synthesize texture $c'$ and displacement $d'$ without the artifacts seen in the projection, while keeping many of the details of the capture.}
    \label{fig:Sarah_AVH}
\end{figure*}

The use of immersive media is rapidly increasing in many industries, for example in education \& training, gaming, and multimedia, but also for therapeutic purposes, e.g., psychotherapy or rehabilitation. The digital modeling of humans holds enormous potential for the development of innovative applications in extended reality (XR), as human models can be used as natural and intuitive interfaces. Modeling and rendering virtual humans in a photo-realistic way is extremely important, as humans are very sensitive to the complex appearance of human bodies and their faces. Further, in XR environments, interactivity plays an important role. The impact and applicability of human models increases significantly with the quality of natural characteristics that can be integrated into the model.

To get towards photorealism in the rendering, the recorded virtual human should appear in a high level of detail. This is possible by leveraging free viewpoint video, a technology where a person is recorded in a studio with dozens of high-resolution video cameras. The multi-camera recording is processed into a sequence of textured dynamic meshes. This sequence allows rendering the recorded performance in XR from different viewpoints in a high level of detail. A major drawback is these mesh sequences cannot be freely animated, as they are confined by their mesh topology. Thus, interactivity is very limited.

Animatability and interactivity can be introduced by leveraging the high flexibility of SMPL \cite{loper2015smpl}. The use of SMPL is a well-proven approach to represent the animation of digital humans but offers low geometric detail even when fitted to real data. We propose to learn pose-dependent appearance and geometry directly from high quality dynamic mesh sequences capturing human performances in the SMPL UV-space in order to facilitate real-time animation.

In this work, we present a novel approach to learning animatable representations of virtual humans from free-viewpoint video data. Our method requires capturing the subject's 3D geometry and texture in high resolution in different poses. The captured data is used to learn a representation of a digital human avatar, which can be rendered in novel poses using standard computer graphics techniques. More specifically, we fit SMPL parameters to the data and learn pose-dependent texture and geometry from the original data in the SMPL uv space. For this, we project the texture from the captured meshes to the posed SMPL mesh and represent the geometry as displacement maps between the model and the scan. Thereby, we transfer the details of the capture to the SMPL uv space in a pose-dependent way and train a model to reproduce these details for novel poses. The original capture and our learned animatable model are contrasted in Figure \ref{fig:teaser}.

The resulting standard computer graphics model (a textured mesh and displacement map) can directly be used in standard virtual and augmented reality (VR/AR) applications without the need for a special rendering code. Our models offer high geometric detail and photo-realistic texture, boosting the immersiveness of XR applications with virtual humans, e.g. in VR education scenarios with a detail-rich teacher model.

We introduce a new representation of animatable virtual humans based on SMPL with learned pose-dependent details trained on performance capture data. We contribute an automatic pipeline to:
\begin{itemize}
\item find correspondences between fitted SMPL meshes and performance capture scans,
\item project textures from the scans onto the SMPL meshes
\item store geometric offsets from the correspondences in displacement maps.
\end{itemize}
We propose a network architecture that learns to synthesize texture and displacement from a given pose. Our approach results in animatable models with detail-rich texture and geometry that can be rendered into novel poses and can be directly integrated into standard XR applications.

\section{Related Work}

Creating virtual humans is an active area of research. Multiple different approaches exist, that yield rendered images of humans with varying levels of detail, depending on the model that was chosen to represent the character. Classical computer graphics methods are often extended with trained neural networks (as in this work). Recently, implicit models have become popular, which represent the characters or scenes with signed-distance functions (SDFs) or Neural Radiance Fields (NeRF). The simplicity of the problem modelling of these methods is very attractive, but their results often have limited resolution. The ray-based rendering methods of NeRFs are quite expensive to compute, which often makes real-time rendering untenable. Furthermore, the integration of ray-based rendering schemes is not straightforward in traditional graphics pipelines that are used to build XR experiences, such as Unity.

In this section, we look at different approaches to virtual human representation, and how they compare to our ideas: SMPLpix \cite{prokudin2020smplpix} builds a neural rendering pipeline, where colors attached to SMPL vertices are fed into a neural rasterizer that produces the final image output. Surface Motion Capture Animation Synthesis \cite{boukhayma2018surface} focuses on creating motion graphs, the appearance is modelled as Eigen textures, decomposing the texture sequences using linear PCA. We build upon this idea by learning the feature cube of textures, which is decoded by a network into the final texture.

DEMEA \cite{tretschk2019demea} encodes a mesh using graph convolutions on a mesh hierarchy, but ignores appearance. Tex2Shape \cite{alldieck2019tex2shape} builds full human body geometry from a single image. Deep4D \cite{regateiro2021deep4d} decodes a skeletal motion into geometry and texture. This is similar to our idea, but we chose to represent geometry in the uv space, to achieve higher geometric detail in the results.

Several approaches have been taken to construct dynamic humans with NeRFs \cite{anerf} \cite{dnerf} \cite{humannerf} \cite{hypernerf} \cite{structnerf}. Most similar to this work may be Surface-Aligned NeRFs \cite{SANeRF}, which work with a correspondence matching approach similar to this work. But their feature space is fixed for the model, and the rendering is orders of magnitude off of being real-time capable. NeRF rendering can be sped up by storing features in hash tables and reducing network sizes \cite{mueller2022instant}. This approach is starting to be applied to animate humans \cite{InstantAvatar}, but resolutions (512x512) and frame rates (15 fps) are insufficient for real-time XR experiences. HDHumans \cite{hdhumans} provide great quality renderings, but train for 10 days, and rendering a single image takes 12 seconds.

Neural Human Video Rendering \cite{liu2020NeuralHumanRendering} encodes pose-dependent details in normal maps and, similar to this work, synthesized partial dynamic textures, but require a refinement network to produce the final output. Generalizable Neural Performers \cite{cheng2022generalizable} combine ray-casting and sampling from target views. In HumanGAN \cite{humangan}, virtual humans are directly synthesized in 2D space, without requiring 3D models.

When a temporal sequence of textured meshes has been constructed from multi-view cameras, the amount of data storage is reduced by at least an order of magnitude, as the duplicate information from the multiple views is condensed into the textured surface mesh. We argue that this is a good starting point to build real-world applications, compared to the listed works above, which mostly work with the camera inputs.

\section{Learning a pose-dependent representation in uv space}
\begin{figure*}
    \centering
    \includegraphics[width=\textwidth]{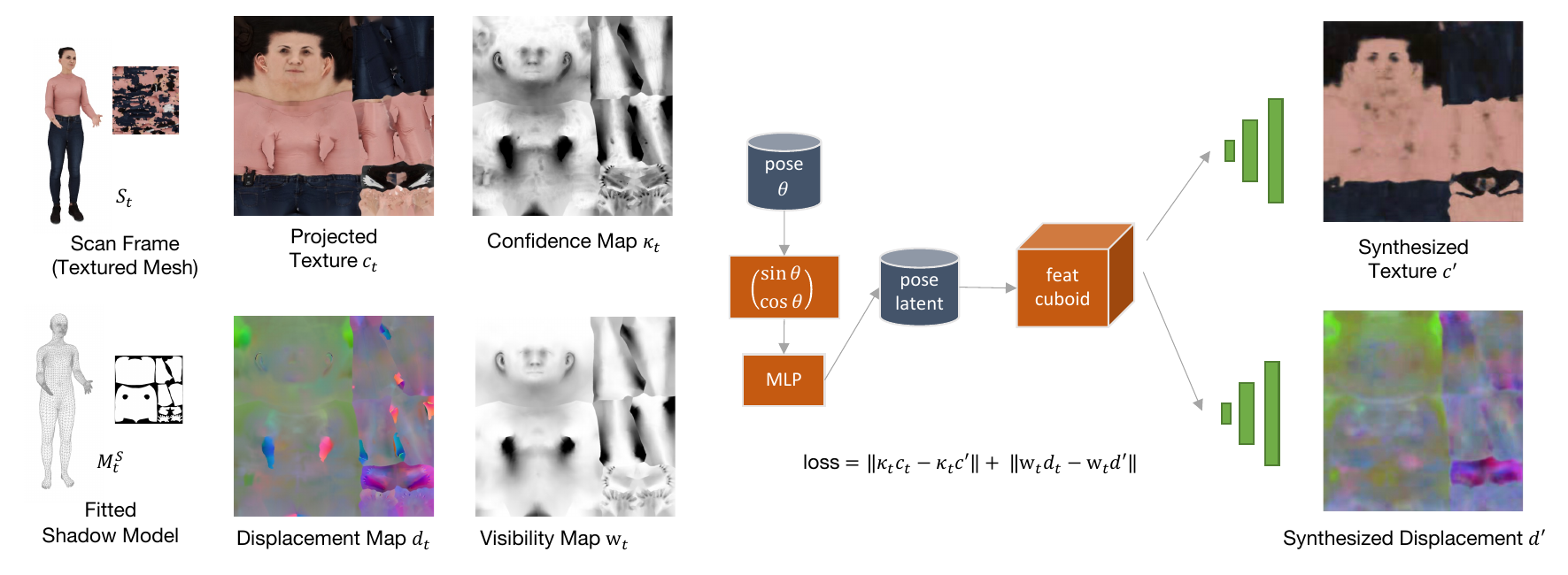}
    \caption{The content from a textured mesh sequence is projected onto a shadow model. The resulting textures, correspondence confidence map, displacement map and surface visibility score are used to train a network to synthesize this data, given a specific pose. In the network, the pose is encoded into a latent space, which selects features from a feature cuboid, which are then successively upsampled into full-size texture and displacement maps.}
    \label{fig:AVH_Arch}
\end{figure*}

In this section, we explain our approach to learn an animatable virtual human representation. We capture a person in a volumetric studio, yielding textured mesh sequences showing the person in a variety of poses with high detail in geometry and texture. A single scan $S_t$ is shown on the left in Figure \ref{fig:Sarah_AVH}). We fit the popular SMPL human body model \cite{loper2015smpl} to the individual frames, yielding SMPL shape parameters per sequence and SMPL pose parameters per frame (see $M_t^S$ in Figure \ref{fig:Sarah_AVH}). The SMPL meshes cannot capture all details from the multi-view video data, but yield a temporally consistent representation. The main idea of our approach is to learn all pose-dependent details in texture and geometry from the performance capture in the SMPL uv space. For this purpose, we further move the surface of the model mesh to match the scan’s surface (see Figure \ref{fig:Sarah_AVH}). We then project the captured textures onto the SMPL mesh and calculate displacement maps between the SMPL mesh surface and the scan data (see Figure \ref{fig:Sarah_AVH} d)). Finally, we train a network that learns a mapping between pose and texture/displacement maps and is able to reproduce them for novel poses with a high level of detail. These are displayed in the same figure after the divider.
 
 In the following sections, we will describe how we map visual details from the textured mesh sequence into the uv space of the template SMPL mesh (shadow mesh), and then train a decoder network to synthesize pose-dependent details in texture and geometry.

\subsection{Input data}

Input to our method is a performance capture of a single person, i.e., a temporal sequence of textured meshes. In our experiments, the characters have been recorded in a volumetric studio with 32 high-definition RGB video cameras, and have been processed into mesh sequences with the pipeline described in \cite{schreer2019capture}.

The mesh topology of the scans changes roughly every second, and with it the texture coordinates of the meshes. Thus, points in neighboring frames that have similar 3D positions may render in the same way, but may map to completely different uv coordinates in texture space. Our idea is to learn a pose-dependent appearance. To allow the network to focus on the appearance changes without having to learn the temporal mapping changes of the uv coordinates, we create a dataset of consistent uv maps by transferring the texture data from the non-consistent textured meshes to a consistent model. For the human representation in this work, we chose the popular SMPL model, which is widely supported and has been proven to support similar use cases well. 

First, we estimate the parameters of the SMPL model to fit it as closely as possible to the scans. Different methods exist to estimate temporally consistent model parameters over performance captures, such as a modified smplify-X \cite{zimmer2023imposing}, PIXIE \cite{PIXIE:2021} and EasyMocap \cite{easymocap}. As the performance captures consist of thousands of frames per take (at 25 fps), we need a method that is fast to compute per frame, robust, and requires little manual oversight and editing. Our data was processed with a variant of EasyMocap, which we extended with an additional loss term representing the average 3D distance between the fitted SMPL shadow mesh surface and the surface of the scan mesh. In the optimization process, we ensure that the SMPL shape parameters $\beta$ for each identity are shared for all recorded frames. This later allows us to make the model solely dependent on the pose of the person, while the shape is fixed.

From the performance capture and model fitting, we thus have a set of global shape parameters $\beta$, and for each frame at time $t$:
\begin{itemize}
    \item $S_t$: the scan, as a textured mesh
    \item global position (translation, rotation, and scale) of SMPL body model
    \item ${\theta}_t$: $(23x3)$ SMPL pose parameters.
\end{itemize}

\subsection{Shadow models: Mapping to SMPL uv-space}

Given joint and shape parameters ${\theta}_t$, $\beta$, as well as the global position for the input data, we can create fitted SMPL meshes $M^S_t$ for each frame. These SMPL meshes are fitted in shape and pose to match the surface of the scan as closely as he SMPL model allows, shadowing it, thus we call the fitted SMPL models \textit{shadow models}. We use the shadow models to create a consistent representation by mapping all details from the performance capture into the shadow uv map, i.e.~the highly detailed textures as well as the geometry details represented as displacement map between the model meshes and the scans. In the following, we describe, how the details from the scans are mapped onto the uv space of the shadow mesh. 

\subsubsection{Pairwise mesh registration}

Due to temporal changes in the geometry of the body and the clothing, a considerable offset between the surfaces of the shadow model mesh $M^S_t$ and the scan $S_t$ is likely to remain. To close this gap, we perform a pairwise mesh registration between the shadow mesh and the scan in order to further move the model surface closer to the scan. This is implemented as an optimization process that moves the vertices of the shadow model $M^S_t$, with a vertex-to-surface loss towards $S_t$ and Laplacian regularization, using the algorithm of \cite{morgenstern2019progressive}. The surface of the registered shadow mesh $M^R_t$ is closer to the surface of the scan $S_t$ than the original model $M^S_t$, improving the quality of the following correspondence finding step.

\subsubsection{Establishing uv correspondences}

The mesh registration process has pulled the surfaces closer by finding correspondences between the model vertices and the scan surface. The goal of the following step is to (i) map the color values from the texture of the scan $S_t$ onto the model mesh and (ii) store the offset between the surfaces in a displacement map in the same space as the texture color. To achieve this, we require correspondences at a higher resolution on the surface at texel level.

There are multiple challenges for finding correct correspondences on texel level: neither mesh is completely wrapping the other, as the two surfaces may intersect. Either geometry may also contain parts that are missing from the other candidate, resulting in surface points without a correct correspondence. The performance capture processing may also have introduced small artifacts in geometry and texture. Given these conditions, the texture baking problem is challenging.

Thus, we implemented the following custom algorithm: First, we create query points in 3D from each valid texel $u^i$ of $M^R_t$ in uv. We chose a resolution of 1024 by 1024 pixels, thus $i=1,\ldots,1024^2$. We map the uv coordinates $u^i$ of the texel to the surface of the mesh, retrieving its coordinates $r_t^i \in \mathbb{R}^3$.

We cast two rays from $r_t^i$ towards the scan $S_t$, one in normal direction of the model surface, and one in negative normal direction. This produces zero, one or two match candidates, when these rays hit the surface of $S_t$. Then we compute whether the normal of the query points in the same hemisphere as the normal at the target surfaces, producing a positive normal match, or not, producing a negative normal match. If all matches have the same normal match polarity, we select the match with the smallest distance. If we have one match with positive normal match polarity and one negative match, we prefer the positive match, unless its distance is more than twice that of the negative match.

\subsubsection{Texture baking and displacement computation}
For all queries of $r_t^i$ that have a match $x_t^i$ on the surface of $S_t$, we now retrieve the color of the scan at this point from its high-resolution texture, and store it as $c_t^i$ in the texture of the shadow model at $u^i$.

The original shadow mesh $M^S_t$ shares its texture space with $M^R_t$, thus for each query point $r_t^i$ we can compute a twin at $s_t^i$: the point on the surface of the shadow mesh with the same texture coordinate. We now store the signed distance $d_t^i = x_t^i - s_t^i \in \mathbb{R}^3$ in a displacement map at $u_i$. We store the displacement map $d$ as floating-point values, to retain the full spatial resolution.

Next, we fill in the texels that are defined in the texture map, but where we could not find matches. The infilling is performed in image space, using an inpainting technique based on the fast marching method \cite{telea2004image} with $radius=3$, implemented in the OpenCV library \cite{opencv_library}. Additionally, we fill in the values of the uv map that were not set, either because no correspondence was found, or because no triangle maps values there, to produce a smooth output image. Finally, not all areas in the texture are defined, some are empty. To reduce artefacts from interpolation at border values of undefined regions when retrieving values from the various textures, we also infill the undefined texel values with the same infilling algorithm as described above.

\subsubsection{Confidence map}
Although the two surfaces have been brought together as close as possible, there will remain parts of the mesh surfaces where the shadow mesh cannot find a true correspondence in the scan, and will find a false match. When training a network on this data, we want it to focus on the correct matches, and ignore false ones. To record an estimation of the correspondence match quality, we compute a confidence map as follows. At each texel, we compute a confidence score, which is a multiplication of four scores covering different aspects of the problem:

First, we consider the visibility of the surface. The color information on the scan surface can only be correct if the texel has been seen by the cameras in the volumetric capture studio. However, we have no information about the cameras and their position in space. Instead, we estimate how well evenly spaced virtual camera positions would capture each surface point.

We compute the visibility $v_t^i$ of scan texels, by self-intersecting with the scan. We cast 64 evenly distributed rays over the hemisphere above the surface around the surface normal. Then, we count how many of the rays self-intersect with the scan and divide by the number of samples. A lower number of self-intersection implies a higher visibility, thus a higher confidence, that the point was seen by one or multiple cameras in the studio: $v = 0.0$ denotes that all samples self-intersected, that the surface lies within the model. $v = 1.0$ denotes that none of the queries found intersections, and the surface points outward. We compute this visibility value for the centroid of each face of the mesh, then we average the values from the face onto its vertices. To get a value for each texel, we use the barycentric interpolation from the values at the vertices.

With this visibility map, we can identify which parts of the scan have texture values that are likely to be incorrect, and should not be use as training examples. Equally important is to identify the inner structure of the model mesh: these inner structures should not find a match on the outside surface of the scan, to not be pulled out from inside the body to its surface. Thus, we compute $w_t^i$, by sampling in the same way over the surface of the registered model mesh.

A high distance between the scan and the model is another aspect that implies a low confidence in the match. We compute the inverse normalized scan distance $\delta_t^i$ by clipping the distance values from the correspondence match to $\frac{1}{50} th$ of the diagonal of the fit extents, and inverting and normalizing the value so that $0$ describes a high distance, and $1$ no distance between the surfaces.

As the last component, we compute a score for the match of the normals, ${nms}_t$, by taking the dot product between the normal of the query point on the surface of the registered mesh $n_t^i$ and the normal of the corresponding point on the scan surface $m_t^i$. We scale them so that a perfect match delivers a value of 1, and opposite normals show a value of 0:
\begin{equation}
    {nms}_t^i = \frac{(n_t^i \cdot m_t^i) + 1}{2}
\end{equation}

The final confidence value $\kappa$ combines the visibility of the points on the scan (the query target, $v_t^i$) and the registered mesh (the query source, $w_t^i$), the distance of the match $\delta_t$, and the normal surface alignment between source and target ${nms}_t$ for the frame at time $t$. Full confidence in the match is denoted by $\kappa = 1$, while $\kappa = 0$ points to the correspondence search having failed.

\begin{equation}
  {\kappa}_t^i = v_t^i * w_t^i * \delta_t^i * {nms}_t^i
\end{equation}

\subsection{Training a generative model for virtual characters}

Given the baked textures and displacement maps in uv texture space of the SMPL model, we can build a network that learns to synthesize these textures and displacements, given the fitted pose that was used to generate the model geometry. An overview of the whole synthesis process is given in Figure \ref{fig:AVH_Arch}.

The network takes the $(23 x 3)$ joint angles $\theta_t$ as the input. To provide continuity of values at $-\pi,\pi$, we encode each angle value $\alpha$ as $(sin(\alpha), cos(\alpha))$. This encoded pose of $138 = 23 * 3 * 2$ values is fed into a small MLP with a single hidden layer of size 256, to map the pose into a latent pose space of size 1024. The latent pose then joins a number of learned 2D feature layers together, forming the initial feature cuboid of size (32x32x1024). The content of the feature maps is learned at training time.

The feature cuboid is convolved with a double convolution with a kernel of size $(3x3)$, followed by a batch norm and ReLU activation, similar to the U-Net architecture \cite{ronneberger2015u}. Then the layer is upsampled with a transposed convolution of kernel size 2 and stride 2, doubling resolution and halving the feature layers. This is continued until the resolution of $1024^2 px$ is reached. A final layer convolves the remaining feature dimension into $RGB=3$ layers. The color values are fed through a final \emph{tanh} activation function to map them to the $[0, 1]$ range.

Replacing the transposed convolution for the upsampling step with a bilinear upsampling followed by a 2D convolution with kernel size 1 to halve the number of layers was found to produce slightly sharper results. To allow the widest compatibility (which becomes helpful when building live demos), the bilinear upsampling was dropped, as it is not supported by popular runtimes (see discussion in section \ref{sec:rendering}).

This autodecoder structure is duplicated, once for the texture synthesis, and once for the displacement map synthesis. The two networks start from the same feature cuboid, but their convolution weights differ. The size of the latent pose code, the number of initial feature layers and their initial size can be adjusted to produce either small networks that compute fast but may lack some detail in the output, or highly detailed networks that take more computational budget. The effect of the various sizes is investigated in section \ref{sec:arch_params}.

For both texture and displacement map, we use the mean square error between ground truth and the result from the network. Each stage that contributes to generating the training data – performance capture, semantic body model fitting, texture projection from correspondences – may introduce visual artifacts, as for example some parts of the body cannot be seen by any camera for a given pose. To avoid having the network learn from incorrect data, we mask the RGB loss of the texture map with the confidence map ${\kappa}_t$ computed alongside the projected texture. As the confidence map contains a term that encodes the inverse distance, we cannot use it to mask the displacement loss. Instead, we use just the visibility term of registered mesh $w_t^i$, so that inner structures (such as the armpit) of the model will not get pulled to the outer surface. We further remove displacement outliers (from mismatches in the correspondence finding algorithm), where points on the scan are more than 5 cm away from the surface of the registered model mesh in any axis. This limit was determined by surveying a histogram of displacement values, and holds well for all sequences processed for this work. As the displacement is computed and synthesized for the posed mesh, the actual displacement values may be much higher, as long as the pairwise mesh registration algorithm was able to pull the mesh onto the scan within 5 cm of each axis.

The final training loss is the sum of the masked texture and displacement losses.

\subsection{Training pose selection}

\begin{figure}
    \centering
    \includegraphics[width=\linewidth]{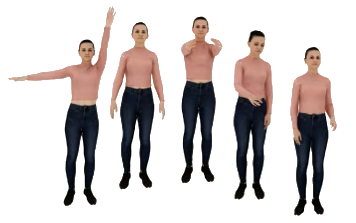}
    \caption{Selecting performance capture frames by pose variance}
    \label{fig:Sarah_Poses}
\end{figure}

Longer captured sequences can contain frames where the person is not fully reconstructed. This may occur when the person moves over the limit of the area that is captured by the multi-camera system. As these broken frames contain no useful texture or geometry data, we discard them if their $\overline{{\kappa}_t} < \frac{130}{255}$.

A short performance capture of 5 minutes will contain 7500 frames (at a frame rate of 25 fps). In many of these frames, the person recorded may be standing around in a default pose, waiting for instructions. To reduce the training time of the network, it is useful to remove frames showing the same pose over and over again. To sample $f$ interesting frames with high variance in the pose (with e.g., $f=100$) from our sample set of around 10,000 frames per character, we implement a selection algorithm that clusters all poses into $f$ k-means clusters of the cosines of the joint angle values. For each cluster, we select the frame that is closest to the cluster's center. This provides a selection of poses with high variance from the training set, while ignoring very similar frames, which would not provide additional benefit to the training data. The first few samples of this sampling mechanism of the Sarah dataset with more than 10,000 frames can be seen in Figure \ref{fig:Sarah_Poses}.

\subsection{Rendering}
\label{sec:rendering}

Once a network has been trained, we can use it to synthesize the animatable virtual human in new poses. The application or experience will request a character for an SMPL pose $\theta$, which is the input to the network. Running the inference with the trained model parameters, the network produces a texture $c'$ and displacement map $d'$ for the given pose. 
\begin{figure}[H]
    \centering
    \includegraphics[width=\linewidth]{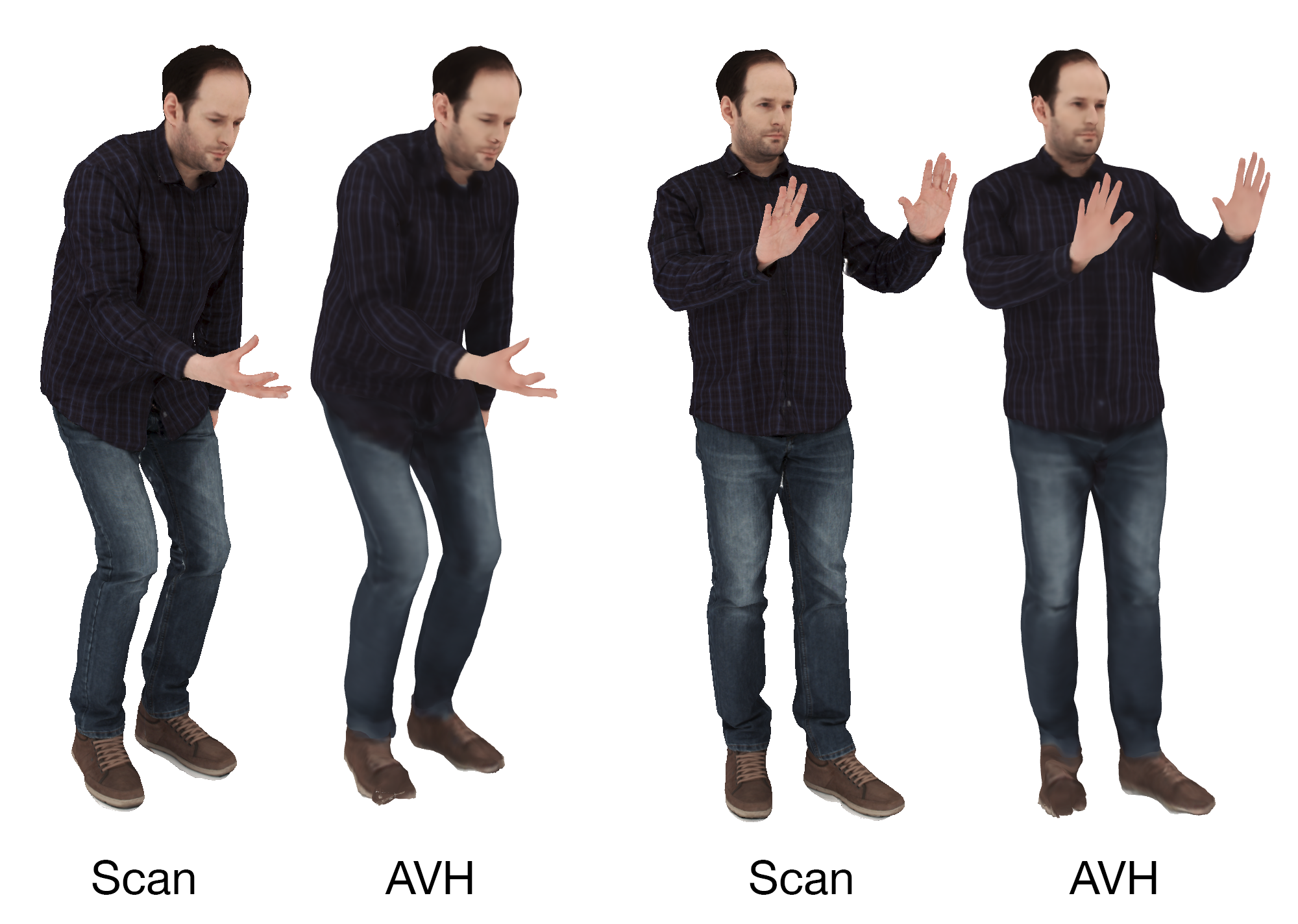}
    \caption{Performance capture frames versus animatable virtual human (AVH) in a grabbing (left two images) and a stop position (right to images).}
    \label{fig:vvf_vs_ahv}
\end{figure}
We apply the pose $\theta$ to build the SMPL mesh $M'$ in the same shape $\beta$ that was used while training the model. The texture $c'$ is then applied to the model. In our offline experiments, we apply the displacement map by subdividing the faces of the SMPL mesh twice and then moving the resulting vertices by the values retrieved from the $d'$. When integrating the network in a live demo, the results should be rendered with a displacement shader for performance reasons.

To enable integration in XR applications, we designed the network to allow exporting it in an interchangeable format. The network can be exported from PyTorch \cite{NEURIPS2019_9015} (which we use for training) in the ONNX format \cite{bai2019} and used for inference via the Barracuda plugin in Unity scenes. As we have chosen to use the standard SMPL model which does not represent finger joints individually, finger shape is not well-represented in the displacement maps. It improves the visual rendering quality to mask out the finger values in the displacement maps.

\section{Results}

\subsection{Qualitative Evaluation}

We train an animatable human model for each actor with $f=100$ frames selected by their pose variance from the roughly 11,000 frames recorded with the actors Sarah and Johnny, respectively. Both actors perform a variety of different actions to ensure pose variety for building models for interactive use cases. Each actor was recorded in four different takes focusing on specific movements:
\begin{itemize}[noitemsep,topsep=0pt]
    \item upper body movement, arm, and head rotations,   %
    \item lower body movement and walking,                %
    \item grabbing, pointing and waving,                  %
    \item talking, including hand gestures.               %
\end{itemize}
We disregard how the character's facial expressions are rendered, as they are not dependent on the pose of the body, and thus should be animated using face-specific algorithms, such as \cite{paier2020interactive}.

In Figure \ref{fig:vvf_vs_ahv}, we compare a recorded performance capture with our rendered animatable virtual human (AVH). We show that our animatable model is able to exactly reproduce each of the shown poses, a grabbing pose and a stop pose. Most notably, our approach to use a displacement map enables our AVH model to render creases and edges of the clothing correctly, e.g., at the edge between shirt and jeans. Furthermore, the displacement map ensures that the head shape matches the corresponding volumetric video. Although our pose dependent texture is not as sharp as in the performance capture, our model provides animatability, while the original capture is fixed at recording time. We still achieve high detail such as facial hair, color changes in the jeans as well as the fine lines of the check shirt. A shortcoming of the underlying body model is that it includes toes, which in our case lead to artifacts as our actors wear shoes. In the supplementary video, we demonstrate how the model behaves under animation, comparing one character to the recorded sequence and showing the other character synthesized in novel poses, that were unseen during training.

\subsection{Metrics and Ablations}
\label{sec:arch_params}

We train our models on two Ampere A100 GPUs with 40 GB of RAM. We run each experiment (listed in Table \ref{table:model_size}) for 500 epochs, with $f=100$ training frames and 10 validation frames. Depending on the selected features and thus the model size, the training time varies between 30-50 min. We use the Adam optimizer \cite{kingma2014adam} with default $\beta=(0.9,0.999)$. Learning rate $lr=0.00131$ and batch size $b=8$ were found by tuning before starting further experiments. We use exponential learning rate decay to speed up convergence with $\gamma=0.99$ determined empirically. %

Following the learning rate experiments, we conducted experiments on feature space size. Here, we focused on the feature cuboid size and the latent pose space. Table \ref{table:model_size} shows the tested network configurations. Alongside the feature cuboid size ($fc$), and latent pose size ($latent$), we recorded the resulting number of parameters ($params$), the final validation loss ($loss$), and the training time ($time$). Lastly, we also measured the inference frame rate ($fps$, in frames per second) for each model on a single A100 GPU. We found that a larger feature space leads to better training performance but does not improve the validation loss and thus fails to improve the generalization performance. In terms of speed ($fps$), all models are feasible to run in real-time; smaller models tend to increase the frame rate. As the final validation loss is lowest for $fc=32$ and $latent=1024$, we continue using these parameters. In our final experiment, we investigated the influence of the hidden layer size in the MLP, setting it to 128, 256, and 512 respectively. Results showed that the hidden layer size does not influence the overall performance significantly; we kept it at 256.

\begin{table}[]
\caption{Feature cuboid size ($fc$) vs. latent pose space ($latent$), best configuration highlighted for each metric}\label{table:model_size}
\begin{tabular}{@{}lrrccc@{}}
\toprule
$fc$      & $latent$    & $params$     & $loss$ ($\cdot 10^{-3}$) & $time$ (min) & $fps$ \\ \cmidrule{1-6} 
16      &    $512$    & $4.8$ M  & $2.56$           &  $\textbf{32}$    & $97$ \\ %
16      &   $2048$    & $73.8$ M & $2.66 $          &  $47$             & $45$ \\\cmidrule{1-6}%
32      &   $1024$    & $19.5$ M & $\textbf{2.37}$  &  $45$             & $49$ \\\cmidrule{1-6} %
64      &   $128$     &  $\textbf{877}$ K & $3.02$  &  $\textbf{32}$    & $\textbf{103}$ \\ %
64      &   $512$     &  $6.8$ M & $2.46$           &  $43$             & $52$ \\ \cmidrule{1-6} %
128     &   $256$     &  $5.4$ M & $2.59$           & $42$              & $57$ \\\bottomrule %
\end{tabular}
\end{table}

\section{Conclusion, Limitations \& Future Work}

We have presented a method for building animatable virtual humans from free-viewpoint video recordings, by baking textures from the capture onto shadow meshes, and then training a neural network to synthesize texture and displacement maps.

We have evaluated our work with a dataset recorded for educational XR applications. We have demonstrated that an animatable character can be automatically created from a few minutes of performance capture, without requiring any intervention by a human artist or designer.

The resulting animatable virtual human can be synthesized from new poses at interactive frame rates. The textured meshes with a displacement that are synthesized can be rendered in a common computer graphics pipeline.

Our approach yield highly realistic animatable models. Still, some limitations remain: The AVH is based on the topology of the SMPL model. Large deviations from the model topology (the person holding props, or garments like skirts) are not directly supported in such an approach. However, we demonstrate the synthesis of pose-dependent displacement maps in this work. Some NeRF-based human models learn static features in an SMPL texture space, which is then used during their ray-based implicit rendering method. It could be an opportunity to leverage our method to build a pose-dependent feature space for NeRF rendering. Also, methods that build a neural re-rendering on top of a rendered SMPL model could benefit from our approach similarly. We are excited to follow up this work by applying either idea.

The training time of less than an hour allows it to be well-integrated into a performance capture pipeline, which usually has a much longer runtime to produce the data. The results of our approach could be improved in future by ensuring that the network produces similar values at both sides of a texture borders. The network size may be reduced by sharing early layers in the displacement and texture synthesis. It would be worth investigating how to disentangle pose and time space,

\acknowledgments{This work has partly been funded by the European Union (SPIRIT, grant no. 101070672), the German Federal Ministry of Education and Research (VoluProf, grant no. 16SV8705), the German Federal Ministry for Economic Affairs and Climate Action (ToHyVe, grant no. 01MT22002A), and the Fraunhofer Society (NeuroHum).}

\bibliographystyle{abbrv}
\bibliography{AVH}

\end{document}